\pdfoutput=1
\documentclass{llncs}

\usepackage{amsfonts,amsmath}

\usepackage{graphicx}
\usepackage{mathptmx}
\usepackage{tikz-qtree}
\usepackage{filecontents}
\usepackage{csvsimple}
\usepackage{amsfonts,amsmath}
\usepackage{subfig} 
\usepackage{url}
\usepackage{verbatim}
\usepackage{color}
\usepackage{amsmath} 
\usepackage{proof}

\renewcommand{\phi}{\varphi}

\input prolog.tex

\definecolor{lgray}{gray}{0.95}
\definecolor{lblue}{rgb}{0.90,0.90,1.00}
\definecolor{lyellow}{rgb}{1.00,1.00,0.70}

\usepackage{listings}
\newtheorem{ex}{Example}

\lstnewenvironment{codex}
    {\lstset{}%
      \csname lst@SetFirstLabel\endcsname}
    {\csname lst@SaveFirstLabel\endcsname}
\newcommand{\BI}[0]{\begin{itemize}}
\newcommand{\EI}[0]{\end{itemize}}
\newcommand{\I}[0]{\item}
\newcommand{\BE}[0]{\begin{enumerate}}
\newcommand{\EE}[0]{\end{enumerate}}

\newcommand{\BX}[0]{\begin{ex}}
\newcommand{\EX}[0]{\end{ex}}
\newcommand{\BV}[0]{\begin{verbatim}}
\newcommand{\EV}[0]{\end{verbatim}}

\newcommand{\BF}[0]{\begin{filecontents*}{data.csv}}

\newcommand{\BQ}[0]{\color{blue}\begin{quote}}
\newcommand{\EQ}[0]{\end{quote}\color{black}}

\def \bscale1 {0.25}
\def \bscale {0.25}

\newcommand{\FIG}[4]{
\begin{figure}[htbp]
\centering
{\includegraphics[scale=#3]{#4}}
\caption{#2}
\label{#1}
\end{figure}
}

\begin{document}

\title{ 
Full Automation of Goal-driven LLM Dialog Threads with And-Or Recursors and Refiner Oracles}

\author{{\bf Paul Tarau}}

\institute{
   {University of North Texas}\\
   {\em paul.tarau@unt.edu}
}

\maketitle

\begin{abstract}
We automate deep step-by step reasoning in an LLM dialog thread by recursively exploring alternatives (OR-nodes) and expanding details (AND-nodes) up to a given depth. Starting from a single succinct task-specific initiator we steer the automated dialog thread to stay focussed on the task by synthesizing a prompt that summarizes the depth-first steps taken so far. 

Our algorithm is derived from a simple recursive descent implementation
of a Horn Clause interpreter, except that we accommodate our logic engine to fit the natural language reasoning patterns LLMs have been trained on. Semantic similarity to ground-truth facts or oracle advice from another LLM instance is used to restrict the search space and validate the traces of justification steps returned as answers. At the end, the unique minimal model of a generated Horn Clause program collects the results of the reasoning process.

As applications, we sketch implementations of consequence predictions, causal explanations,  recommendation systems and topic-focussed exploration of scientific literature.

\end{abstract}


\section{Introduction}

Interaction with today's high-end LLMs like ChatGPT, GPT-4 \cite{gpt3,gptrl} and Bard \cite{bard2023} allows the patient and prompt-savvy user to steer the interaction toward fulfillment of a well-specified information seeking goal. The resulting dialog thread can be labor intensive and assumes solid prompt engineering skills to keep the LLM focussed on the task while digging as deep as needed into details.

This raises the obvious question: can we  get back the simplicity of a one-shot  query and automatically manage the navigation in the results space of the dialog thread?

We start by sketching here an answer to this question. Clearly, we need first an elaboration or refinement process that reduces a given task to a sequence of subtasks. We call this conjunctive elaboration into subtasks an {\em AND-step}.
Next, we will need a dual, disjunctive elaboration, as a generation of alternative ways to make progress on the task. We call this an {\em OR-step}.
To advance into more detail we can rely on a recursive process that alternates these two steps up to the desired depth.

{\em This brings us to the key topic of this paper: an algorithm that extracts a salient set of answers, by zooming into the desired level of detail, from a single, succinct human prompt. To this end, we  automate step-by step reasoning in an LLM dialog thread to explore recursively alternatives (OR-steps) and expand details (AND-steps) up to a given depth.}

Our approach will  follow closely the SLD-resolution algorithm for pure Horn Clause logic \cite{sld,kowalski76}. Restriction to Horn Clauses is motivated by the fact that  LLMs  are genuinely ``constructive'' and known not to be comfortable with negation \cite{qunatanot,blanconot}, limiting one's interest in either classical negation nor negation-as-failure under a closed-world assumption as present in ASP systems \cite{GelfondL88,asp} or in Prolog \cite{swi}.


This makes the use of a conventional logic programming language unnecessary as Python's coroutining generators  are expressive enough for succinctly implementing a simplified SLD-resolution algorithm \cite{iclp21}.
Another departure from logic programming as we know it, is that we will need to "unformalize" the underlying logic to more easily interoperate with the LLMs.
In fact, LLMs do have a limited ability to generate correct logic forms of simple sentences \cite{gopal23}. But  their training is based mostly on completion of natural language sentences and they are more in their element with the  reasoning steps humans express in natural language.

{\em Thus, instead of trying to force LLMs to use logic formalisms they are not yet comfortable with, we
accommodate our logic engine to fit natural language reasoning, goal driven planning, task decomposition and association patterns with minimal task-specific prompt engineering.}

This brings us to the key features of our approach:
\BI
\I SLD-resolution's clause selection via unification is replaced by LLM-driven dynamic
clause head creation with an option of focusing by proximity of embeddings to ground truth facts
\I as dialog units are sentences, the underlying logic is propositional
\I client-side management (via the API) of the LLM's memory is based on the equivalent of a {\em goal stack}, as used in logic-programming implementations like \cite{iclp17,iclp21}, and a {\em goal trace} recording our steps on the current search path
\I instead of variable bindings, answers are traces of justification steps clearly explaining where they are derived from
\I when their depth-limit is reached, the items on the goal stack are interpreted as "abducibles" that might be 
hypothetically assumed and then checked against "integrity constraints" \cite{abductive,abd_lp}.

\I our depth-bounded refinement steps support compilation of the dialog threads to a
Horn Clause program to be explored with logic programming solvers

\I modular, task specific, customizable prompt engineering primitives are aggregated together for ``AND-step'' and ``OR-step'' prompts

\I normalized semantic similarity measures of embeddings can be made available when generating probabilistic logic programs
\I sentences in authoritative documents or collections seen as "ground-truth facts" can be used to select abducibles via semantic similarity or advice of an LLM-based oracle
\EI

{\em Overall, our approach exploits synergies between structured prompt engineering, logic-guided recursion over LLM queries and semantic search in an embeddings vector store.}

Applications are built  by customizing prompts, LLM models and recursion level, resulting in automatically generated  detailed, ``hallucination-free''  answers, crisper and more accurate than what one can obtain after lengthy interactions with conventional search engines or Chat-GPT style dialog agents.

Among potential applications we will overview the following:
\BI
\I consequence predictions and causal explanations with full justification traces
\I recommendation closely focussed on initial preference seed
\I actionable step-by-step advice on practical ``how to repair'' problems 
\I topic-focussed scientific literature keyphrase and key concept generation
\EI

The rest of the paper is organized as follows:
Section \ref{sys} sketches the architecture of our implemented system.
Section \ref{inter} introduces Interactors -- designed by aggregating base classes needed for interacting with LLM APIs.
Section \ref{rec} describes Recursors -- our programs steering the LLMs dialog threads while focusing on the task at hand over multiple levels of nested OR-steps and AND-steps.
Section \ref{appr} describes Refiners,  specializations of Recursors checking against ground-truth facts using semantic distances to abducible facts as well as several LLM-based oracle agents. 
Section \ref{model} describe our propositional Horn Clause model generator that extracts the set of true facts inferred from the logic program generated by our Recursors and Refiners.
Section \ref{appl} shows how task-specific applications are built simply by adjusting the AND-step, OR-step and oracle prompts.
Section \ref{disc} discusses variations on the main theme of the paper and possible future extensions.
Section \ref{rel} discusses related work and section \ref{conc} concludes the paper.

\section{System Architecture}\label{sys}
\FIG{arch}{System Architecture}{0.075}{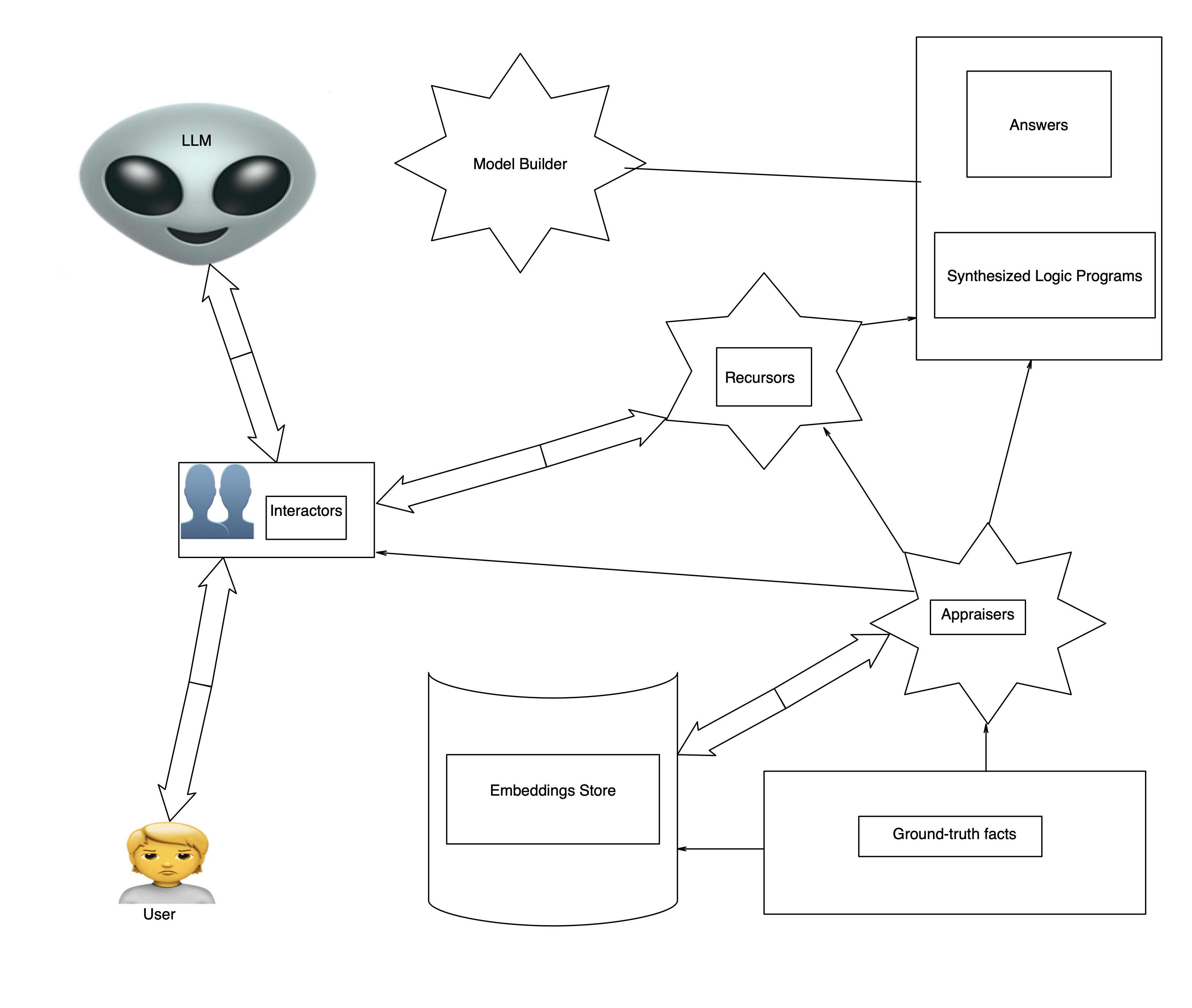}

We start with a quick overview of an implemented system architecture and a sketch of its execution flows, as shown in Figure \ref{arch}.

Starting from a succinct prompt (typically a nominal phrase describing the task) an Interactor will call the LLM via its API, driven by a Recursor that analyzes the LLM responses and activates new LLM queries  as it proceeds to refine the information received up to a given depth. Refiners are Recursor subclasses that rely on semantic search in an Embeddings Store containing ground-truth facts as well as on oracles implemented as specialized Interactors that ask the LLM for advice on deciding the truth of, or the rating of hypotheses. Besides returning a stream of answers, Recursors and Refiners compile their reasoning steps to a propositional Horn Clause program available for inspection by the user or subject for execution and analysis with logic programming tools (in particular, with our Model builder -- a fast Propositional Horn Clause theorem prover).

\section{Interactors}\label{inter}

Setting up the interaction mechanism with LLMs via an API is a multi-faceted process involving several orthogonal aspects.

We will overview here our modular, multiple-inheritance based separation of concerns for interacting with LLMs centered at this time around OpenAI's {\tt gpt-4} and {\tt gpt-3.5-turbo} models \cite{gpt3,gptrl}, but ready to accommodate also smaller footprint, locally running LLMs like LLaMA \cite{touvron2023llama}.

An interactor will be put together by inheriting from the base classes Prompter, Tuner, Cacher and Tracker or a selection of their task specific subclasses.
This gives significant (statically-checked) flexibility in customizing them.

\subsection{Tuner}
Tuner is the base class for setting LLM API parameters (including models). Inheriting from a selection of well balanced Tuners known to work well on specific tasks simplifies application development.
\subsection{Cacher}
Cacher is the base class for our persistence mechanism avoiding repeated LLM API calls on reruns.
It also automates (via Python meta-programming) the caching of attributes in human readable json form.
Its {\tt persist} method automates saving to disk all built-in datatype and collection attributes of an Interactor.
Its {\tt resume} method restores a saved state if the cache file exists. Finally, its {\tt clear} method
erases it.

\subsection{Prompter}\label{pro}

Prompters use patterns expressed as Python dictionaries from which substitution of  
 {\tt \$}-variables with data provided at various recursion levels will 
generate actual prompts to be sent to the LLMs.

Here is an overview of our Prompters' key features and use cases:
\BI
\I a Prompter is a base class for aggregated, task specific OR and AND prompt generators or decision Oracles
\I on top of them we build a collection of task specific  parametric prompt templates
\I AND and OR prompt templates for a given task are designed together to facilitate their experimental fine-tuning
\I prompt templates instantiate one-shot instructions to the LLM that enforce focussed, succinct answers
\I possible post-processors (algorithmic or  implemented as ``verifier'' LLM oracles) can be used
to discard answers when the LLM disobeys the instructions either in 
requested syntactic form or in content.
\EI

We will show next a few prompt template examples. Note as the LLMs (in this case GPT-3.5 and GPT-4) and their APIs evolve, minor edits might be needed to adjust them to the changes. 

\BX
AND-OR prompt patterns for causal reasoning

\begin{codex}
causal_prompter = dict(
    name='causal',
    
    and_p="""We need causal explanations in this context: "$context"
       Generate 3-5 explanations of 2-4 words each for the causes of "$g".
       Itemize your answer, one reason for "$g" per line.
       No explanations needed, just the 2-4 words noun phrase, 
       nothing else.
       Your answer should not contain ":" or "Cause".
        """,
        
    or_p="""We need causal explanations in this context: "$context"
       Generate 2-3 alternative explanations citing facts that might
       cause "$g".
       Itemize your answer, one noun phrase per line.
       No explanations needed, just the noun phrase, nothing else.
       Avoid starting your sentence with the word "Alternative".
       Your answer should not contain ":" .
       Your answer should avoid the words "Causes" and "causes" ."""
)
\end{codex}
Note that the {\tt \$context} and current goal {\tt \$g} parameters will
specialize the pattern for each of the uses of its {\tt and\_p} and {\tt or\_p} components
in the recursive descent process. Note also the ``petty'' avoidance remarks in the
prompt that we had to use to ensure that the answer returned by the LLM matches
the expected output structure, given that after parsing, it has to provide the inputs of the next
recursive step.
\EX

\BX
Oracle pattern used to filter hypotheses generated by a Recursor
\begin{codex}
decision_prompter = dict(
    name='oracle',
    
    decider_p="""
    You play the role of an oracle that decides if "$g" is relevant for 
    our interest in "$context".
    Your answer should be "True" or "False" expressing agreement or 
    disagreement with the relevance of "$g".
    """
)
\end{codex}
The pattern is used to decide about adequacy of a given subtask or alternative in a given context.
With a similar rater oracle we request ratings on a scale of 100 if we want to generate a probabilistic logic program to be analyzed with tools like Problog \cite{de2007problog,deepproblog1}.
\EX

\subsection{Tracker}
Tracker is the base class  managing API messages, contexts and API costs. It also ensures that
answers to questions already answered by the LLM are cached and reused to save costs and ensure
full determinism and replicability. It also handles the on-demand migrations from an Interactor's
short-term memory to its long-term memory. While the short-term memory is kept small enough to fit in the LLMs message size, both memories are dictionaries used to retrieve available cached answers.
As a special case, Trackers also enable spilling of the full content of the short-term memory
to the long-term memory when a fresh dialog thread is needed for a change of topic or focus.

\subsection{Talker}

Talker is the base class managing the overall interaction with the LLM. It assumes all other components have been already inherited by the Interactor that inherits from it. Its high level {\tt ask} method
encapsulates the details of applying the appropriate prompt template to a given question,
activates mechanisms to trim the context to a size acceptable to the LLM, activates
conversion of the content of the short-term memory to the message format the LLM expects.
It also activates possible application-specific post-processing of the LLM's answers and 
manages the retrying of the completion request if the API is temporarily unresponsive.

\section{Recursors}\label{rec}
Recursors implement the central idea of this paper: automatically focusing a dialog thread with an LLM, while exploring a given topic in depth.

\subsection{The recursive descent}
Starting from a succinct initiator goal, the Recursor performs a recursive descent guided by task specific Prompters.  At each step, the OR-prompter asks the LLM to generate alternative ways to fulfill the current goal.
Then,  for each alternative, the LLM is asked to expand it to a sequence of task specific subgoals, guided by the AND-prompter.
 The AND and the OR prompter templates are activated with information about the current context and the current goal.
The context is a linearization of a chronologically ordered trace that accumulates the previously expanded goals. The presence of this context, serving as the {\em short term dialog memory} of the otherwise stateless LLM API, steers the generative process to stay focussed on the task.

At a given step, the effect of the OR-prompter expanding the head $h$ in a series of alternatives $a_1,\ldots,a_n$ can be described as a set of binary Horn Clauses of the form:\\\\
\begin{math}
h~:-~a_1.\\
h~:-~a_2.\\
\ldots\\
h~:-~a_m.\\
\end{math}\\
On the other other hand, the result of an AND-prompter can be described as a set of Horn clauses of the form:\\\\
\begin{math}
h~:-~b_1,~b_2,~\ldots,~b_n.\\
\end{math}

When a depth limit is reached, the remaining unexplored goals on the goal stack are considered as {\em abducibles} \cite{abductive,abd_lp}, i.e., hypotheses to be assumed until integrity constraints might invalidate them, a process that, like in Logic Programming, results in backtracking to explore other possibilities. Should some of them fail, the presence of the OR-nodes at each recursive step ensures that plenty of choices remain available, despite possible failures. 

A simple way to select abducibles is to check the semantic distance of their embeddings to embeddings in a set of {\em ground-truth facts}.
For efficiency, a few nearest neighbors of each abducible are fetched from the vector embeddings store (see subsection \ref{store})  and their average distance to the ground truth is used to decide if the abducible is assumed as a hypothesis.
A summary of the sentences extracted from the ground-truth facts can be used as an explanation supporting the abducible. This can be seen as an instant constraints-driven filtering operation that results in eagerly omitting the assumption of the irrelevant abducibles as hypotheses.

Besides returning a stream of answers, we
    also generate a propositional Horn Clause program to be further
    explored with logic programming tools.

At the end, a minimal model \cite{kowalski76} of the remaining rules can be obtained with a SAT solver, although our implementation prefers a fast direct algorithm (see section \ref{model}), given that Horn Clause satisfiability is polynomial \cite{dowling}.

\subsection{The Unfolder}

Our implementation of the depth limited recursive descent encapsulates the unfolding of AND-nodes and OR-nodes. An Unfolder instance contains two Interactor Agents, one for each node type, initialized with their jointly designed prompter dictionary described in section \ref{pro}. The agents are activated with the {\tt ask\_and} and {\tt ask\_or} methods and are also responsible for persisting past LLM interactions in appropriately named unique disk caches.

\subsection{The AndOrExplorer}
The AndOrExplorer implements its recursive descent by relying directly on the Python-stack and emulating Prolog's backtracking via Python's {\tt yield}-based coroutining mechanism. It returns the trace of expanded goals  (and invented clause heads)  for each successful ``proof step'', assuming all facts at the depth limit as abducibles.

The clause invention step is sketched by the following Python code snippet:
\begin{code}
    def new_clause(self, g, trace, topgoal):
        or_context = to_context(trace, topgoal)
        hs = self.unf.ask_or(g, or_context)
        and_context = to_context((g, trace), topgoal)
        for h in hs:
            bs = self.unf.ask_and(h, and_context)  # invent their bodies
            yield (h, bs)
\end{code}
The {\tt or\_context} is built from the generic OR-pattern instantiated to the specifics of the step in the {\tt trace} of the goals expanded so far on this resolution branch. Note that {\tt topgoal}  is also passed to the context builder to help focus on the original goal that has started the recursive descent. It is responsible for the generation of the list of clause heads {\tt hs}. For each clause head {\tt h} in {\tt hs}, a clause body {\tt bs} is generated by the AND-node prompter. Finally each clause is yielded as a pair  {\tt (h, bs)}. 

Besides its {\tt resume},  {\tt persist} and {\tt costs} methods the {\tt AndOrExplorer} defines also an {\tt apprise} method meant to be overridden by its refiner subclasses.

\subsection{The GoalStacker}

 The GoalStacker  implements a similar recursive descent steering the LLM to
    expand conjunctively or disjunctively while
    staying focussed via a controlled history of
    explored goals and unfolding new ones.
    It uses an explicit goal stack that gives access to
    continuations and depth markers for
    each of the goals allowing to return
    richer and deeper answers.
 Similarly to the AndOrExplorer's {\tt apprise} method, this recursor implements an overridable {\tt filter} method that
 enables {\em decider} and {\em rater} oracles (see section \ref{appr}) to reject abducibles (and implicitly their possible consequences), that are likely to diverge from their current context and original goal.
 
\subsection{The Logic Programming connection}

The recursive descent algorithm is implemented as a generator of answers (traces of steps included) to the initiator goal, in a way similar to Prolog's SLD-resolution algorithm operating on Horn Clauses. In fact, its Python implementation has been derived as a simplification of of the Natlog \cite{iclp21} Horn Clause interpreter, where clause selection via the unification algorithms is replaced by synthesis of a clause head by an OR-node. Then, given the clause head, {the body of the clause is generated by the LLM as an an AND-node expansion of the synthesized clause head}.

Instead of the {\tt true} or {\tt fail} answer generated by a Prolog system running the propositional Horn Clause program, the complete trace of goals generated by the LLM and ``solved'' by our recursor is returned as an ``explanation'' of the ``reasoning steps'' taken in the process.
In fact, the resulting Horn Clauses are also "compiled" on the fly to a Prolog program that could be independently explored with a Prolog, Datalog or  ASP system. However, given the presence of loops (as the LLM might come back in the recursive process to things it has already seen), we use instead of Prolog a low polynomial-time model builder that is insensitive to the presence of loops \cite{dowling}.

\section{Refiners}\label{appr}
Refiners are extensions of Recursors evaluating AND-nodes and OR-nodes against ground-truth facts in the embeddings store or via {\em decision} or {\em rating} LLM-oracles
(see subsection \ref{ora}).

In the first case, normalized semantic distances between embeddings of a goal hitting a depth limit 
and ground-truth facts are used. If close enough to a ground-truth fact, the ``abducible'' goal becomes  a clause head and the ground-truth fact becomes the body of a newly generated clause. If not close enough to any ground-truth fact, the goal becomes the head of a clause marked for failure when compiled to the Prolog program by having as body the atom {\tt fail}. 

Alternatively, abducibles can be evaluated by an oracle -- another LLM instance that judges
 their relevance to the task in the current context.

\subsection{The Embeddings Vector Store}\label{store}
We use FAISS \cite{faiss}, a scalable vector store supporting efficiently ground-truth collections of fact
embeddings provided by Sentence Transformer models \cite{sbert} or embeddings obtained via the LLM's API. 
The ground-truth facts store is then used to find the K nearest neighbors of a given query
and also to return  Euclidean or normalized dot-product distances, the latter usable as probabilities to decide what hypotheses
can be assumed during the recursive descent as ``abducible'' facts, subject   to filtering via 
evaluation of integrity constraints. Scalability of FAISS to billion-size stores supports the use of a  knowledge base possibly derived from a large document collection.

\subsection{Filtering with semantic distance to ground-truth facts}

 By using the ground-truth facts in our embeddings store the simplest way to apprise if a given goal is ``on topic'' is to compute its semantic distance to its nearest neighbor in the store, as shown in the following code snippet:
\begin{code}
    def appraise(self, g, _trace, _topgoal):
        rates, neighbors = self.store.qa(g, top_k=1)
        rate, neighbor = float(rates[0]), neighbors[0]
        return rate > self.threshold:
\end{code}
The method {\tt qa} that queries the store passes the goal {\tt g} and the request for a single nearest neighbor {\tt top\_k}.

A more elaborate technique relies on $k$ nearest neighbors fetched from the store that would collectively
 ``champion'' the goal if their (weighted) average semantic distance to the goal is below a threshold, fixed in advance or dynamically computed or machine-learned form past appraisals.

The mechanism can also be extended to continuously check for staying close in terms of semantic distance to the ground-truth facts.

Alternatively, the semantic distances (interpreted as probabilities) can be used to annotate clause heads as part of a probabilistic logic program to be evaluated later.

Oracles can also be used to implement {\em continual appraising}: at each step they can check reasonable closeness to ground truth and task at hand. In particular,  they can mark confidence level for each rated step and then select overall highest only.

\subsection{Refining decisions with LLM-based oracles}\label{ora}
In the absence of a set of ground-truth facts relevant to a given initiator goal, the LLM itself
can be asked to make True/False decisions or generate ratings.

\subsubsection{The LLM-based True/False Decider}

The following code snippet delegates the steering to focus on a given context (in this case the initiator goal that has started the recursive descent). In this case, the {\tt apprise} method instantiates the oracle prompt pattern shown in subsection \ref{pro}.
\begin{code}
    def appraise(self, g, trace, topgoal):
        advice = just_ask(self.oracle, g=g, context=topgoal)
        return 'True.' == advice
\end{code}

More elaborate refiners can use the depth-first path {\tt trace} or an LLM-generated summary thereof
instead of {\tt topgoal}. 

\subsubsection{The LLM-based Rater}

The Rater queries the LLM asking for a score on the 0 to 100 scale that is next converted into a probability. Like the Decider oracle, it uses   the goal at hand and its context. For both oracles, the prompter can be configured to ask for an explanation sentence or paragraph  to be adopted as ground-truth in case of favorable True/False decision or high enough confidence level.

\section{The Model builder: a  Propositional Horn Clause Satisfiability solver}\label{model}

It is not unusual to have loops in the propositional Horn Clause program connecting the LLM generated items by our recursors and refiners. As that would create problems with Prolog's depth-first execution model, we implement a simple low-polynomial complexity propositional satisfiability checker and model builder along the lines of \cite{dowling}.

The model builder works by propagating truth from facts to rules until a fix point is reached. Given a Horn Clause $h:-b_1,b_2,...,b_n$, when all $b_i$ are known to be true (i.e., in the model), $h$ is also added to the model. If integrity constraints (Horn clauses of the from $false:-b_1,b_2,...,b_n$) have also been generated by the oracle agents monitoring our refiners, in the advent that all $b_1,b_2,...,b_n$ end up in the model,  $b_1,b_2,...,b_n$ implying $false$ signals a contradiction and thus unsatisfiability of the Horn formula associated to the  generated program.  However as the items generated by our recursive process are not necessarily expressing logically connected facts (e.g., they might be just semantic similarity driven associations), turning on or off this draconian discarding of the model is left as an option for the application developer.
Also, the application developer can chose to stop as soon as  a proof of the original goal emerges, in a way similar to goal-driven ASP-solvers like \cite{casp}, irrespectively to unrelated contradictions elsewhere in the program.

\section{Applications}\label{appl}

A good hint on deciding which recursor or refiner is the most appropriate for developing an application, is the closeness of its atomic steps to processes of human problem solving that are  similar to logic inferences, e.g., by sharing a similar underlying boolean algebra, lattice or preordered set structure.
Besides causal reasoning or consequence prediction most goal-oriented tasks (e.g., planning) fit this structure. It is also good to be aware that when exploring the causes or the consequences of an initial state of the world, technological development, military, political or judicial decision, it is likely that the LLM will generate a richer model than if it explores names of movies, books or songs in a recommender system, where titles are often overlapping semantically with unrelated embeddings.
In the former, restricting the model with a stricter oracle can even out spurious facts. In the latter, being aware that the LLMs will work better on very well known movies or books than when asking for recommendations similar to a relatively new or niche product, can guide the scope of the application.

When developing an application that, starting from a keyphrase of a scientific paper or the name of a scientific domain, would generate highly relevant concepts for the domain, an oracle set up to filter out less specific concepts from en encompassing more general domain can help with the return of more salient results. In this case a second oracle, filtering out generic methodological boilerplate concepts, shared by virtually all scientific domains will be also useful to give more focussed results.
 
In the case of requesting expert advice on practical common tasks (e.g., actionable step-by-step advice on how to repair something), the oracle can filter out advice to contact the manufacturer or seek the advice of an expert nearby, when the point is to  receive the actual steps need to solve the problem.
This can also be implemented by a set of negative ground facts for which a refiner can try to maximize  average semantic distances or a post-processor that rejects choices containing words or keyphrases in a blacklist.
 
Another kind of application of significant practical value is to use a set of generated models to benchmark newcomer LLMs' performance against established best in their class like, at the time of writing this paper, OpenAI's GPT-4. This can be achieved with something as simple as the Jaccard distance between the inferred models at a given depth and it can be fine-tuned to the specific task the LLMs are planned to be used for. 
Related to this, when transferring from strong LLMs like GPT-4 to weaker ones, it can be useful to train the Reinforcement Learning loop rewards to be based on how many hits the weaker LLM gets when recursing on a relatively small collection of ``critically important topics'' automatically collected from the stronger LLM, thus providing an novel, potentially very effective transfer learning mechanism.
 
We refer to the Appendix for illustrations of execution traces, snippets of  code and models.
The generated Prolog programs, models and execution traces are available at:\\
 \url{https://github.com/ptarau/recursors}.

\section{Discussion on Limitations and Variations on the Theme}\label{disc}

\subsection{Limitations}

We will next overview some of the limitations we have experienced when testing (the current implementation of) our recursors and refiners when on several target applications.

First, let us note that recursors are obviously not needed when one-shot detailed descriptions of common processes (e.g., cooking recipes, tell a joke, write a haiku) are available directly from the LLM.
They are also  unnecessary for help from the LLM to write a news story, a bio, an essay, a resume or an add, where interactive fine-tuning of the LLM's in-context learning by the user is a clear requirement.

The mapping between the recursively generated items and propositional logic does not apply  equally to all tasks and the inference steps work differently when the LLM is used simply as an associative-memory connecting concepts interesting as brainstorming incentives to humans, but not meant to be logically focussed on a dominant topic or task.

Indicators like semantic similarity are not relevant for recommendation systems over items consisting of titles of movies, books or songs where their distributional semantics is dominated by the more common uses of the title's actual word phrases. 

Strictly enforcing integrity constraints generated by oracles looking at local contexts often results, when propagated through the inference process, in unsatisfiability (and thus an empty model). Note however that this limitation  can be alleviated by accepting a partial model supporting the initial goal, even if the resulting logic program might be inconsistent, an option supported by our model generator described in section \ref{model}.

There's increased sensitivity to prompts during deeper recursive thought-to-thought steps. In this case, careful prompt engineering is needed as recursors can easily induce ``butterfly effects'' - small variations in wording of the prompt can drastically change the resulting model.

\subsection{Future Work Directions}
Once the key idea of the paper for steering LLMs to generate a stream of items focussed on the initiator goal is implemented, several ``variations on the theme'' can be tried out relatively easily by overriding methods in Prompters, Recursors and Refiners.

\subsubsection{Granularity Refiners}
 At a higher granularity one can work with sentences/statements instead of noun phrases, sometime a more natural match to the underlying propositional logic.
 
 At a lower granularity, one might want to use SVO triplets that LLMs are quite good at decomposing a sentence to. The generated SVO triplets can then serve as building blocks for a Description Logics or Datalog programs.

 \subsubsection{Question generators}
 
Answers to LLM-generated how+wh-questions for a given goal can be used as expansions to new goals simply by re-engineering our AND-OR Prompters.

\subsubsection{Generalizers} 
 
Inductive Logic Programming techniques can be used to generalize the resulting propositional or triplet clauses by sharing common SVO fragments, possibly in combination with a Prolog-rules describing the ground-truth background knowledge.

 \subsubsection{Diversifiers and Harmonizers}
 
 To further restrict unwanted ``hallucinatory'' generation twists {\em diversifiers} for OR-nodes and {\em harmonizers} for AND-nodes can be expressed as additional integrity constraints.
A diversifier will work by ensuring no two OR-nodes are too close semantically while a harmonizer will ensure that no two AND-nodes are too far semantically. Both could be implemented with help of semantic distances in the embeddings store.

\subsubsection{Extend Implementation Techniques}

 One can implement recursors with with bare completion-only LLMs (e.g., GPT3) as their usual question-answering fine-tunings (e.g., ChatGPT) can be emulated with minor prompt engineering efforts.
 
 
To limit the scope of decider and rater oracles, on can preprocess the ground-truth facts  into FAISS-supported k-means clusters, and restrict oracle search to the cluster closest to the initiator goal.

\subsubsection{Extend the power of the underlying logic language}

It is possible to use SAT-based ASP solvers \cite{asp} or Prolog-based CASP systems  \cite{casp} to take advantage of failed LLM returns rejected by our oracles to enhance the ability of the LLMs to reason with negative information in a principled way as well as with negative ground-truth facts meant to avoid extending into semantically close but distinct domains during the recursive descent.

\section{Related Work}\label{rel}

The major disruption brought by the often near-human quality of generative AI \cite{gpt3,gptrl,touvron2023llama} is quickly changing the landscape of query-driven information retrieval, moving the emphasis from traditional search engines 
to human-friendly dialog threads. However, the effectiveness of actionable information extraction is often hindered by the slower partner in this interaction -- the human that needs to understand, evaluate and validate each step.
In this context, our work emphasizes the full automation of this retrieval process, starting from a succinct query term. Thus, we are back to the one-shot simplicity of "short question by human $\rightarrow$ arbitrarily deep, elaborated answer from the AI", steered to stay focussed on the actual query. As a  side effect of this automation, our approach  preempts most of the usual problems with hallucinations, lack of factuality and bias that LLMs are often blamed for.

Our recursive descent algorithm shares with work on ``Chain of Thought'' prompting of LLMs \cite{chainofthought,ling2023deductive} and with step by step \cite{lightman2023lets} refinement of the dialog threads the goal of extracting more accurate information from the interaction. However our process aims to fully automate the dialog thread while also ensuring validation of the results with help of ground-truth watching oracles and independent LLM-based agents. Our approach shares with tools like LangChain \cite{langchain} the idea of piping together multiple instances of LLMs, computational units, prompt templates and custom agents, except that we fully automate the process without the need to manually stitch together the components.

By contrast to ``neuro-symbolic'' AI \cite{neurosym}, where the neural architecture is closely intermixed with symbolic steps, in our approach the neural processing is encapsulated in the LLMs and accessed via a declarative, high-level API. This reduces the semantic gap between the neural and symbolic sides as their communication happens at a much higher, fully automated and directly explainable level.

In \cite{gopal23} LLMs are cleverly used to generate Prolog code snippets with an enhanced CASP \cite{casp} semantics. This allows successfully hand-building useful applications like a conversational AutoConcierge bot, recommending local restaurants. By contrast, our method is generic, and ``no code'' applications consist simply in queries with possible minor prompt engineering, as we adapt logic programming to think in terms familiar to LLMs, rather than adapting LLMs to generate application specific code snippets.

Also within Logic Programming, in relation with probabilistic approaches \cite{de2007problog}, our abducibles acquire probabilities from normalized vector distances to ground truths, usable to automate generation of probabilistic logic programs, thus sharing objectives with  more directly neuro-symbolic approaches like \cite{deepproblog1}. Finally, we share with \cite{tplp22} the idea to use of a custom logic solver (an Intuitionistic Propositional Theorem prover, in that case) to synthesize abducibles (the Propositional Horn Clause model generator in our case).

\section{Conclusion}\label{conc}

We have automated deep step-by step reasoning in LLM dialog threads up to a given depth, by recursively descending from a single succinct task-specific phrase, while staying focussed on the task at hand.
In the process, we have made LLMs function as de facto logic programming engines that mimic Horn Clause resolution. However, instead of trying to parse sentences in into logic formulas, we have accommodated our logic engine to fit the natural language reasoning patterns LLMs have been trained on. 
Semantic similarity to ground-truth facts and oracle advice from another LLM instance has been used to restrict the search space and validate the traces of justification steps returned as answers. 
This has resulted in focussed, controllable output, enabling deep investigations into details of specific scientific domains as well as expert-level causal reasoning or consequence predictions. As such, our approach streamlines  key use cases of LLMs as focussed information seeking tools and enables practical application back-ends simply by customizing prompt templates.


\bibliographystyle{splncs}

\bibliography{theory,tarau,ml,proglang,biblio,ref}

\section*{Appendix}
\BX
Complete short run to depth=1 for initiator goal = ``Logic programming'':
\begin{codex}
TRACE:
Logic Programming
Symbolic reasoning
Symbolic representation learning

TRACE:
Logic Programming
Symbolic reasoning
Knowledge-based systems

TRACE:
Logic Programming
Symbolic reasoning
Automated theorem proving

...

TRACE:
Logic Programming
Predicate calculus
Herbrand universe
\end{codex}

\begin{codex}
MODEL: 19 facts, alphabetically ordered 

Automated theorem proving
Expert system development
First-order logic
First-order logic inference
Herbrand universe
Inference engine mechanisms
Knowledge representation formalisms
Knowledge-based systems
Logic Programming               <=== initiator goal
Predicate calculus
Production rule systems
Quantifier elimination
Resolution principle
Rule-based systems
Semantic networks
Semantic reasoning algorithms
Skolemization process
Symbolic reasoning
Symbolic representation learning
\end{codex}
Note that the trace consists of hitting the limit depth of 1 after one OR-node and expansion and picking each of the nodes of the subsequent AND-mode expansion.
\EX

\BX
The generated Prolog program for initiator goal = ``Logic programming'':
\begin{codex}
% RULES:
'Logic Programming' :-
    'Symbolic reasoning'.
'Logic Programming' :-
    'Rule-based systems'.
'Logic Programming' :-
    'Predicate calculus'.
'Symbolic reasoning' :-
    'Symbolic representation learning',
    'Knowledge-based systems',
    'Automated theorem proving',
    'First-order logic inference',
    'Semantic networks'.
'Rule-based systems' :-
    'Knowledge representation formalisms',
    'Inference engine mechanisms',
    'Production rule systems',
    'Expert system development',
    'Semantic reasoning algorithms'.
'Predicate calculus' :-
    'Quantifier elimination',
    'First-order logic',
    'Skolemization process',
    'Resolution principle',
    'Herbrand universe'.
\end{codex}

\begin{codex}
% FACTS
'Symbolic representation learning'.
'Knowledge-based systems'.
'Automated theorem proving'.
'First-order logic inference'.
'Semantic networks'.
'Knowledge representation formalisms'.
'Inference engine mechanisms'.
'Production rule systems'.
'Expert system development'.
'Semantic reasoning algorithms'.
'Quantifier elimination'.
'First-order logic'.
'Skolemization process'.
'Resolution principle'.
'Herbrand universe'.
\end{codex}
Note that in this case no oracle has been used to trim the model, by using the basic AndOrRecursor class as the generator, with a prompt exploring a scientific topic. All the abducibles (goals hit a level 1) have been assumed true.
\EX

\BX
Prompt used in the previous run, focussed on following concepts in a scientific research field:
\begin{codex}
sci_prompter = dict(
    name='sci',
    and_p="""The task we are exploring is: "$context"
        Generate 3-5 noun phrases of 2-4 words each that occur as 
        keyphrases only in scientific papers bout "$g".
        Itemize your answer, one noun phrase per line.
        No explanations needed, just the noun phrase, nothing else.
        """,
    or_p="""The topic we are exploring is: "$context"
        Generate 2-3 noun phrases describing details of "$g".
        Itemize your answer, one noun phrase per line.
        No explanations needed, just the noun phrase, nothing else.
        """
)
\end{codex}
\EX
Next we will illustrate some uses of Refiners, this time for deeper (but also more selective)  exploration of other domains. Given the length of the traces and models, we will abbreviate them focusing on the most salient 
fragments\footnote{Full traces of 
executions are available at \url{https://github.com/ptarau/recursors}}.

\BX
Trace of a Refiner using the Decider oracle to validate causal explanations on the topic ``Biased AI''.
\begin{codex}
!! ADVICE for: Lack of diverse data True.
...
!!! ADVICE for: Intellectual Property False.

TRACE:
Biased AI
Lack of diverse data
Limited data sources
Lack of incentives for companies to share data
Privacy Concerns
...
!!! ADVICE for: Incomplete data False.
!!! ADVICE for: Limited representation of minority groups in 
    training data True.
!!! ADVICE for: Lack of diversity True.
...
TRACE:
Biased AI
Human bias in data collection
Stereotyping
Limited representation of minority groups in training data
Lack of diversity
...
\end{codex}
\EX

\BX
Snippets from a Horn Clause program generated by a recursor on practical goal ``Repair a flat tire''
\begin{codex}
'Repair a flat tire' :-
    'Replace the flat tire with a spare tire'.
'Repair a flat tire' :-
    'Patch the punctured area of the flat tire'.
'Repair a flat tire' :-
    'Inflate the flat tire and check for leaks, then repair any leaks 
    found'.
'Replace the flat tire with a spare tire' :-
    'Park the vehicle on a flat surface and engage the parking brake',
    'Locate the spare tire and the necessary tools, such as a jack 
    and lug wrench',
    'Use the lug wrench to loosen the lug nuts on the flat tire, but 
    do not remove them yet',
    'Position the jack under the vehicle in the appropriate location 
    and raise the vehicle until the flat tire is off the ground',
    'Remove the lug nuts and the flat tire from the vehicle',
    'Place the spare tire onto the vehicle s wheel base and 
    hand-tighten the lug nuts',
    'Lower the vehicle back to the ground using the jack',
    'Use the lug wrench to tighten the lug nuts in a star pattern 
    until they are snug',
    'Double-check that all lug nuts are tightened properly',
    'Store the flat tire and tools in the trunk or designated area'.
'Patch the punctured area of the flat tire' :-
    'Locate the punctured area of the flat tire',
    'Remove any debris or foreign objects from the punctured area',
    'Roughen the area around the puncture with sandpaper or a wire brush',
    'Apply rubber cement to the roughened area and allow it to dry 
    for a few minutes',
    'Peel the backing off the tire patch and place it over the 
    punctured area, pressing firmly',
    'Use a roller or the edge of a coin to ensure the patch is 
    securely bonded to the tire',
    'Inflate the tire to the recommended pressure and check 
    for any leaks',
    'If there are no leaks, reattach the tire to the vehicle and 
    tighten the lug nuts'.
...
'Inflate the flat tire and check for leaks, then repair any leaks found':-
    'Locate the flat tire and remove it from the vehicle',
    'Place the tire on a flat surface and remove the valve cap',
    'Attach the tire inflator to the valve stem and inflate the tire to 
    the recommended pressure',
    'Use a spray bottle with soapy water to check for leaks around the 
    valve stem and tire bead',
    'If a leak is found, mark the location and deflate the tire',
...
\end{codex}
\EX

\BX
Trace of a Refiner using a Rater oracle with ground truth facts fetched from Wikipedia article on topic ``Expansion of the universe''

\begin{codex}
RATING of "Dark energy" w.r.t "Expansion of the universe"
is 0.593 --> True
RATING of "Dark energy" w.r.t "Expansion of the universe"
is 0.593 --> True

TRACE:
Expansion of the universe
Dark energy
Vacuum energy
Accelerating universe
Dark energy

RATING of "Vacuum energy" w.r.t "Expansion of  the universe" 
is 0.5524 --> True

TRACE:
Expansion of the universe
Dark energy
Vacuum energy
Accelerating universe
Vacuum energy
...
RATING of "Energy density" w.r.t "Expansion of the universe" 
is 0.3581 --> False
RATING of "Quantum fields" w.r.t "Expansion of the universe" 
is 0.5799 --> True

TRACE:
Expansion of the universe
Dark energy
Vacuum energy
Cosmological constant
Quantum fields

RATING of "Zero-point energy" w.r.t "Expansion of the universe" 
is 0.3579 --> False
RATING of "Accelerating universe" w.r.t "Expansion of the universe" 
is 0.7293 --> True
RATING of "Dark energy" w.r.t "Expansion of the universe" 
is 0.593 --> True

TRACE:
Expansion of the universe
Dark energy
Cosmological constant
Accelerating universe
Dark energy
\end{codex}
\EX

\BX
A Decider recommender systems with "The Godfather" movie as initiator.
\begin{codex}
!!! ADVICE for: The Shawshank Redemption True.
!!! ADVICE for: The Dark Knight False.
!!! ADVICE for: Pulp Fiction True.
!!! ADVICE for: Reservoir Dogs False.
!!! ADVICE for: Kill Bill: Vol. 1 False.
!!! ADVICE for: Inglourious Basterds False.
!!! ADVICE for: Fight Club False.
!!! ADVICE for: Pulp Fiction True.
!!! ADVICE for: Reservoir Dogs False.
!!! ADVICE for: Kill Bill: Vol. 1 False.
!!! ADVICE for: Inglourious Basterds False.
!!! ADVICE for: The Dark Knight False.
!!! ADVICE for: Fight Club False.
!!! ADVICE for: The Silence of the Lambs False.
!!! ADVICE for: Pulp Fiction True.
!!! ADVICE for: Reservoir Dogs False.
!!! ADVICE for: Goodfellas True.

TRACE:
The Godfather
The Shawshank Redemption
The Departed
Pulp Fiction
Goodfellas
...
!!! ADVICE for: The Silence of the Lambs False.
!!! ADVICE for: Kill Bill False.
...
!!! ADVICE for: Fight Club False.
!!! ADVICE for: Goodfellas True.
!!! ADVICE for: The Irishman True.
...
!!! ADVICE for: Fight Club False.
!!! ADVICE for: The Dark Knight False.

MODEL: 11 facts 

Goodfellas
Pulp Fiction
Reservoir Dogs
Scarface
The Departed
The Godfather
The Godfather: Part II
The Irishman
The Shawshank Redemption
The Sopranos (TV series)
The Untouchables
\end{codex}
\EX

\BX
Horn Clause program (snippets) for Refiner using a Rating oracle with Wikipedia page as background truth on topic ``Artificial general intelligence'', with rating threshold = 0.35
\begin{codex}
'Artificial general intelligence' :-
    'Cognitive abilities of AGI'.
'Artificial general intelligence' :-
    'Learning algorithms of AGI'.
'Artificial general intelligence' :-
    'Problem-solving capabilities of AGI'.
'Cognitive abilities of AGI' :-
    'Reasoning skills',
    'Learning capacity',
    'Problem-solving ability',
    'Cognitive flexibility',
    'Decision-making process'.
'Reasoning skills' :-
    'Logical deduction abilities'.
'Reasoning skills' :-
    'Pattern recognition skills'.
'Reasoning skills' :-
    'Decision-making capabilities'.
'Logical deduction abilities' :-
    'Propositional reasoning skills',
    'Syllogistic reasoning abilities',
    'Deductive inference capacity',
    'Logical deduction aptitude',
    'Reasoning with uncertainty'.
'Pattern recognition skills' :-
    'Visual pattern recognition',
    'Speech pattern analysis',
    'Object recognition abilities',
    'Facial recognition skills',
    'Pattern matching algorithms'.
'Decision-making capabilities' :-
    'Adaptive decision-making',
    'Probabilistic reasoning',
    'Cognitive flexibility',
    'Multi-criteria decision-making',
    'Heuristic decision-making'.
...
'Cognitive flexibility'.
'Probabilistic reasoning'.
'Reinforcement learning'.
'Working memory'.
'Transfer learning'.
'Domain adaptation'.
'Adaptive decision-making'.
'Ethical decision-making'.
...
\end{codex}
\EX

\end{document}